\documentclass[runningheads]{llncs}

\usepackage{accv}

\usepackage{accvabbrv}

\usepackage{graphicx}
\usepackage{booktabs}

\usepackage{amsmath}
\usepackage{amssymb}
\usepackage{amsfonts}
\usepackage{booktabs}       
\usepackage{multirow, tabularx}
\usepackage{enumitem}
\usepackage[detect-all,per-mode=symbol]{siunitx}
\usepackage[style=english]{csquotes}
\usepackage[all,british]{foreign}

\renewcommand\foreignabbrfont{\normalfont}
\DeclareRobustCommand{\vs}{\xperiodafter{{\foreignabbrfont{vs}}}}

\usepackage[accsupp]{axessibility}  

\usepackage{hyperref}

\usepackage{orcidlink}

\begin{document}

\title{Gravity-guided Contact Dynamics Estimation from 3D Human Motions} 

\titlerunning{GraCE}

\author{
Cuong Le\inst{1}~\orcidlink{0000-0002-8509-4952}, 
Urs Waldmann\inst{2}~\orcidlink{0000-0002-1626-9253}, 
Bastian Wandt\inst{2}~\orcidlink{0009-0002-1203-1093}, 
Mårten Wadenbäck\inst{1}~\orcidlink{0000-0002-0675-2794}
}
\institute{
Linköping University, Sweden. \,\texttt{\{firstname.lastname\}@liu.se}
\and
Independent researcher
}

\maketitle

\begin{abstract}

Ground contact forces acting on the human body, are crucial for biomechanics studies or sport performance analysis.
Prior methods rely on force plates or pressure mats to collect ground contact dynamics, limiting their applicability to carefully controlled settings.
A more scalable solution is to estimate the dynamics directly from motion capture data.
Recent approaches only roughly estimate the ground contact dynamics from the vertical distance between the body and the ground plane, which cannot capture the complex pressure distribution of all contact points.
To this end, we propose GraCE -- \underline{Gra}vity-guided \underline{C}ontact Dynamics \underline{E}stimation, a novel full-body contact dynamics model for human motions using a realistic influence of body mass distribution and gravity.
We use the human's center of gravity to estimate the ground contacts based on its relative distance to the human body.
The applied force on each contact is estimated via the product of predicted contact probabilities and the total exterior force computed from the center of mass trajectory.
We outperform related work on the GroundLink dataset for ground reaction force estimation, and on the MOYO dataset for detailed contact pressure prediction.
The code is available at our \href{https://github.com/cuongle1206/GraCE}{GitHub}.

\end{abstract}
\begin{figure}[t]
    \centering
    \includegraphics[width=0.65\linewidth]{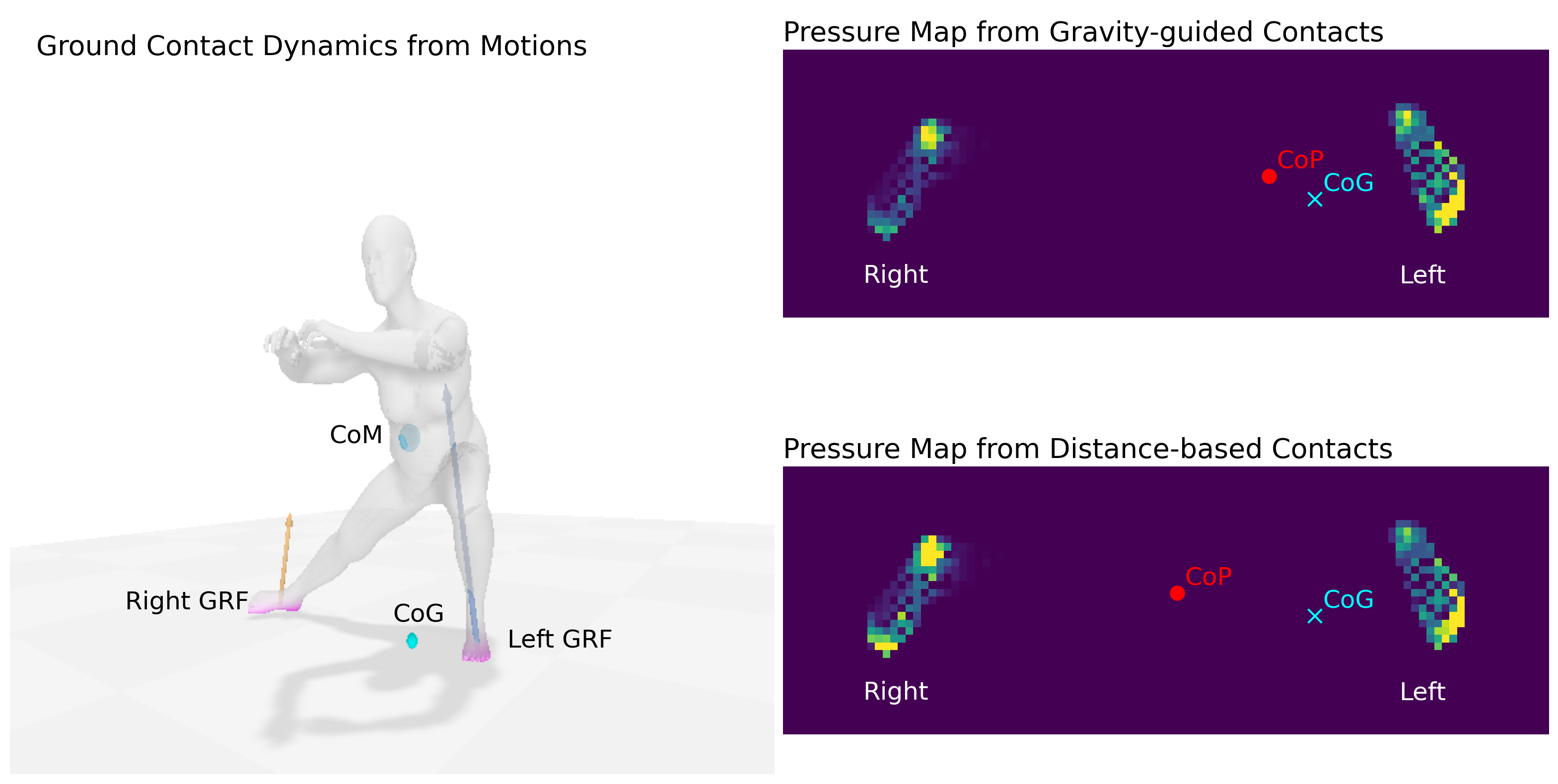}
    \caption{
    Gravity-guided \vs distance-based contacts.
    An example of an ambiguous case during a stretching exercise, where each foot contact is at the same vertical level to the ground, but the reaction force is mainly distributed toward the left foot, as the ground reaction force (GRF) in left contact (\textcolor{NavyBlue}{blue} arrow) is larger than right contact (\textcolor{Orange}{orange} arrow).
    The gravity-guided model imposes the influence of the center of gravity CoG (\textcolor{Cyan}{cyan}) to the contacts, leading to correct pressure estimation with the CoP (\textcolor{Red}{red}) closer to the left foot, unlike the distance-based model that distributes the reactions evenly between the two feet with the CoP in the middle.
    }
    \label{fig:teaser}
\end{figure}

\section{Introduction}
\label{sec:introduction}

Plausible recovery of ground contact dynamics is an essential ingredient for many biomechanics applications such as clinical gait analysis~\cite{silder15_job, weyand00_joap}, sport performance analysis~\cite{bates15_joks, steele12_gap, zhao24_os}, and human movement studies~\cite{seth18_opensim, uchida20_bom}.
Traditionally, contact dynamics are collected from sensor devices in laboratory environments, \ie force plates for ground reaction force and center of pressure measurements.
This laboratory setting restricts biomechanics research to simple motions performed by actors in a small volume.
Motion capture systems, on the other hand, offer a higher volume of data collection and a wider area of application, enabling the actors to freely perform complex movements.
Therefore, recent approaches often estimate ground contact dynamics directly from the captured motions, using the Newton mechanics as the foundation.
The total exterior force that causes the human motion trajectory is often simplified as a residual force acting on the root or the center of mass of the human body~\cite{seth18_opensim, brubaker09_estimcd}.
The total exterior force accounts for all of the reaction forces from the body's contacts with the surroundings, commonly the ground, and produces the human trajectory via an ordinary differential equation (ODE) with the acceleration computed from Newton's second law of motion.
Given a motion, GraCE learns to predict the total contact reaction force that could reconstruct the center of mass (CoM) trajectory of the input motion, using a learnable PD controller with the concept of residual force~\cite{brubaker09_estimcd} and Newton mechanics.

To provide meaningful data for further biomechanics research, the total force must be redistributed to either joint-level or vertex-level contact force~\cite{uchida20_bom}.
Previous methods mainly model contact probability via the vertical distance of the joint/vertex to the ground, with the reaction force scaled accordingly.
These distance-based methods often struggle with subtle movements, where the vertical distance of multiple contacts is the same but the magnitude of reaction force applied to them is different.
\cref{fig:teaser} illustrates an example of a stretching motion, where the feet have the same vertical distance to the ground, but the distribution of the reaction force differs significantly, causing distance-based methods to fail.
To this end, we propose a novel approach that predicts contact probability based on the center of gravity (CoG) for more accurate dynamics estimation in subtle movements.
The human body is modeled via the SMPL+H model, a vertex-based volumetric model that can be controlled via a set of pose and shape vectors \cite{romero17_smplh}.
Inspired from biomechanics literature~\cite{winter95_balance, huec11_eqbody, vielemeyer21_job}, we design a new contact dynamics model for every body vertex based on the distance between the CoG and the projection of the vertex to the ground.
The contact model parameters are learned via a CNN in a fully differentiable pipeline.

Furthermore, since the contact model includes all body vertices, the dynamics can be easily adapted to specific tasks.
Ground reaction forces, similar to force plate data, is obtained by summing all the force from the respective foot vertices; or a ground pressure map, similar to data from pressure mats, can be computed via the summation of forces from all body vertices projected to the area of interest.
GraCE is evaluated with related ground contact models on: 1) the GroundLink dataset~\cite{han23_groundlink} for ground reaction force estimation in comparison to force plate measurements; and 2) the MOYO dataset~\cite{tripathi23_ipman} for full-body pressure estimation with pressure mat data.
In both setups, GraCE outperforms the related methods by a significant increase in contact estimation accuracy.

\section{Related work}
\label{sec:related_work}

\subsection{Physics-based motion capture}
\label{subsec:related_mocap}

Prior work in vision-based human motion capture uses physics constraints, the contact dynamics from the ground plane, to reduce jittery and unnatural poses through either: 1) physics engines for imitation learning~\cite{yuan2021_simpoe, yao2022_cvae, huang2022_neuralmocon, yuan2023_physdiff}; 2) motion optimization objectives~\cite{rempe2020_eccv, xie2021_iccv, gartner2022_trajopt, gartner2022_diffphy}; or 3) acceleration modeling for motion reconstruction~\cite{shimada2021_neurphys, rempe2021_humor, li2022_dnd, le2024_osdcap}.
The exterior reaction in these approaches is often simplified via detections of contact events based on thresholding, which leads to the discontinuities of reaction dynamics and can only be applied to a fixed set of body joints such as feet or hands.
These simplifications result in implausible and discontinuous dynamics capture, which is undesirable for further biomechanics analysis.
We address these limitations by proposing a continuous contact model for all body vertices represented via the SMPL+H model~\cite{romero17_smplh}. 

\subsection{Contact dynamics from motions}
\label{subsec:related_dynamics}

Recent work leverages the power of neural networks by learning to predict contact dynamics using motion captures as input data.
GroundLink~\cite{han23_groundlink} learns to reproduce the force plate measurement of ground reaction forces and center of pressure using a temporal convolutional network, with SMPL motions as inputs.
Phys-GRD~\cite{le2025_physgrd} reinforces the learning of GroundLink with an additional physics-based loss, reinforcing the missing of force plate measurements via simulated signals.
UnderPressure~\cite{mourot22_underpressure}, and the line of research from~\cite{scott20_dynamics,scott22_stability,kraiger25_footformer} learn to predict foot pressures using neural networks and foot sole sensor data as ground truth.
ImDy~\cite{liu25_imdy} addresses the limitations of sensory measurements to laboratory setups by simulating human dynamics via motion imitation with reinforcement learning on the AMASS database~\cite{mahmood19_amass}, and learns to reproduce the simulated dynamics with a transformer network.
These approaches rely on large-scale databases and complex learning architectures for good estimations, and often struggle with unseen complex human movements.

A more robust method for estimating contacts is by explicitly modeling the contact probabilities via the kinematics interactions between the human body and the ground.
Some prior work from~\cite{brubaker09_estimcd} and~\cite{zell17_iccvw} estimates motion exterior dynamics using a continuous contact model based on the vertical distance between human 3D keypoints and the ground.
The contact probability is bounded by a sigmoid function, and the motion dynamics is computed via optimization to minimize the exterior force applied to the body root joint, while satisfying Newton's second law of motion.
To impose more sophisticated contact estimations, recent work utilizes the human volumetric model SMPL as the body representation.
PhysPT~\cite{zhang24_physpt} uses a model based on the vertical distance and velocity, also inspired by~\cite{brubaker09_estimcd}, to estimate the contact probability for a subset of vertices of the SMPL model that frequently make contact with the ground in the AMASS dataset.
IPMAN~\cite{tripathi23_ipman} instead models the full body ground contacts via a combination of exponential and linear functions as the vertex penetrating the ground plane, thus implemented as an objective loss for the 3D human pose estimation.
Also, as pointed out in IPMAN~\cite{tripathi23_ipman}, and other biomechanics literature, the stability of the human body has strong correlation with the distance between the CoG and the center of pressure (CoP).
The closer the CoG to the CoP, the more stable the human poses are~\cite{winter95_balance, huec11_eqbody}.
We find that the captured dynamics, including \eg reaction forces or contact pressures, also contain the corresponding influences from the CoG to every body contact position.

The prior work mainly models the contact probability via the vertical distance of the vertex to the ground and by-pass the stability condition imposed by the CoG.
This leads to incorrect contact dynamics estimations, especially when encountering complex poses with multiple contacts at the same vertical level while the reaction forces are distributed differently between contacts.
We design a new contact model that explicitly considers the CoG influence to every body vertex, quantified by an exponential function with parameters adaptively learned via neural networks.

\section{Method}
\label{sec:method}

\begin{figure}[t]
    \centering
    \includegraphics[width=0.98\linewidth]{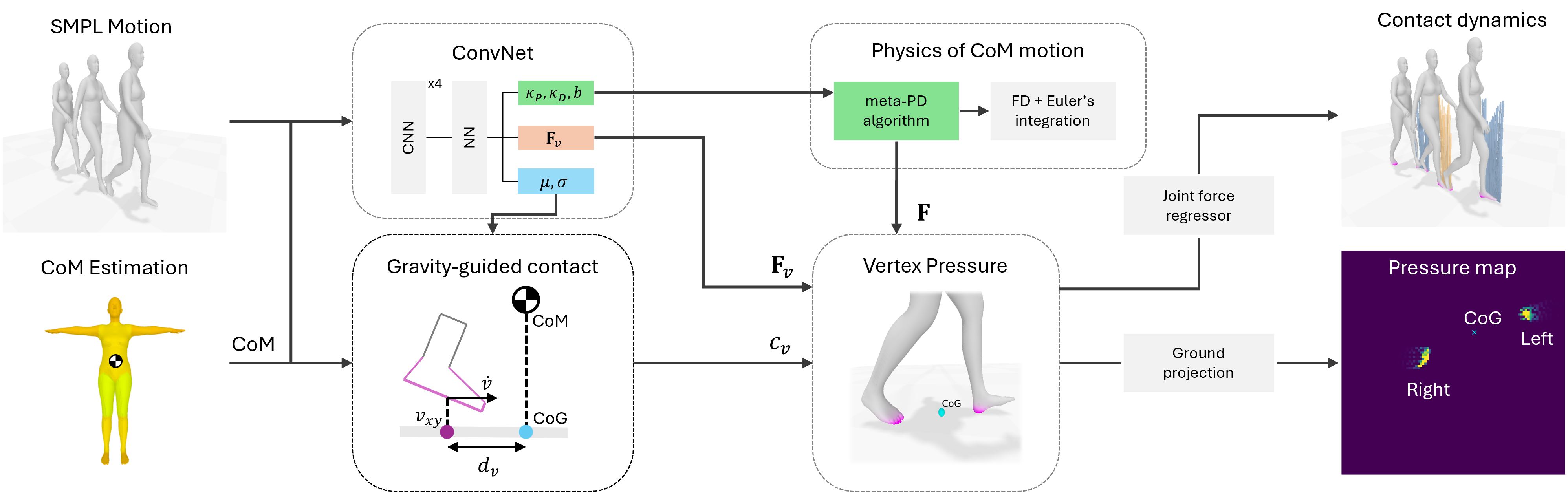}
    \caption{
    GraCE consists of two separate tasks: 1) contact probability estimation for every body vertex, and 2) total exterior force estimation from input kinematics motion.
    The gravity-guided contact probability of each body vertex is computed based on its vertical distance and velocity, scaled by the exponential output from the distance between the vertex to the CoG.
    The magnitude influence from the CoG is determined by the predicted parameters $\mu, \gamma$ from the ConvNet.
    Concurrently, the total exterior force $\mathbf{F}$ is estimated via a PD controller using learned control parameters $\kappa_P, \kappa_D, b$ (also predicted by the ConvNet) to minimize the CoM trajectory reconstruction loss.
    With the assumption of a conservative system, the respective reaction force of each body vertex is the total $\mathbf{F}$ scaled by the softmax outputs.
    }
    \label{fig:pipeline}
\end{figure}

\subsection{Overview}
\label{subsec:overview}

\cref{fig:pipeline} shows the overview of our method.
We first compute a sequence of SMPL meshes $\{\mathbf{V}_t \in \mathbb{R}^{6890\times3}\}^T_{t=1}$ parameterized by the sequence of root translations $\{\mathbf{y}_t \in \mathbb{R}^3\}^T_{t=1}$, body poses $\{\theta_t \in \mathbb{R}^{24\times3}\}^T_{t=1}$ in axis--angle representation, and the shape $\beta \in \mathbb{R}^{10}$.
Then, we estimate the sequence of CoM $\{\mathbf{x}_t \in \mathbb{R}^{3}\}^T_{t=1}$ based on the common mass distribution of the human body from \cite{delava96_mass}.
We define the CoG as the vertical projection of the CoM to the ground~\cite{winter_biomov09}.
In \cref{subsec:exterior}, we show how we estimate the total exterior force $\mathbf{F}$ from the CoM sequence $\{\mathbf{x}_t\}^T_{t=1}$.
\cref{subsec:contact} presents our novel gravity-guided contact model based on the CoG.
\cref{subsec:dynamics} illustrates how the GRF on specific body parts or the detailed ground pressures can be regressed directly from the full-body vertex pressures.

\subsection{Exterior force estimation}
\label{subsec:exterior}

We aim to simulate a new CoM trajectory $\{\mathbf{r}_t\}^T_{t=1}$ to match the original trajectory $\{\mathbf{x}_t\}^T_{t=1}$ via Newtonian mechanics.
At each time step $t$, the total exterior force $\mathbf{F}$ must satisfy the equation of motion: $\mathbf{F} + \mathbf{G} = m\Ddot{\mathbf{r}}_t$, where $\mathbf{G} \in \mathbb{R}^3$ is the constant gravitational force applied to the CoM, $m \in \mathbb{R}$ is the total mass of the person, and $\Ddot{\mathbf{r}}_t \in \mathbb{R}^3$ is the acceleration of the simulated CoM.
Inspired from~\cite{shimada2021_neurphys,li2022_dnd,le2024_osdcap}, the exterior force $\mathbf{F}$ is estimated via the learnable PD controller:
\begin{equation}
    \mathbf{F} = \kappa_P (\mathbf{x}_{t+1} - \mathbf{r}_t) - \kappa_D \dot{\mathbf{r}}_t + b_t,
    \label{eq:fext}
\end{equation}
where $\kappa_P$, $\kappa_D$, and $b$ is control parameters predicted from the ConvNet.
To simulate the new CoM trajectory $\{\mathbf{r}_t\}^T_{t=1}$, we apply the semi-implicit Euler integration scheme:
\begin{equation}
    \begin{split}
        &\dot{\mathbf{r}}_{t+1} = \dot{\mathbf{r}}_t + \Ddot{\mathbf{r}}_t \Delta t, \\
        &\mathbf{r}_{t+1} = \mathbf{r}_t + \dot{\mathbf{r}}_{t+1} \Delta t, \\
    \end{split}
    \label{eq:euler}
\end{equation}
where $\dot{\mathbf{r}}$ is the simulated velocity and $\Delta t$ is the time step between two consecutive frames.
We train the convolutional neural network, ConvNet~\cite{han23_groundlink}, to minimize the mean squared errors between the simulated $\{\mathbf{r}_t\}^T_{t=1}$ and the original CoM trajectory $\{\mathbf{x}_t\}^T_{t=1}$:
\begin{equation}
    \mathcal{L}_{CoM} = \frac{1}{T} \sum^T_{t=1} (\mathbf{r}_t - \mathbf{x}_t)^2 ~.
    \label{eq:L_com}
\end{equation}

The estimated $\mathbf{F}$ is used to explain the residual forces applied to the person's body, not accounted by the internal body joint torques and gravity \cite{brubaker09_estimcd}.
The source of external force mostly comes from the ground plane, via multiple contact points and accumulates at the CoM.

\subsection{Gravity-guided contact}
\label{subsec:contact}

To compute the specific ground dynamics, we first need to find the contact probability of each vertex.
For an SMPL mesh $\mathbf{V}_t \in \mathbb{R}^{6890\times3}$, our contact model for each vertex $\mathbf{v} \in \mathbb{R}^3$ is demonstrated as ($t$ is omitted for better readability):
\begin{equation}
    c_\mathbf{v} = 4\underbrace{\sigma(-\psi_{d} \mathbf{v}_z)}_{\textup{distance term}}~\underbrace{\sigma(-\psi_{v} \Vert\dot{\mathbf{v}}\Vert)}_{\textup{velocity term}}~\underbrace{\exp{\frac{-\Vert \mathbf{v}_{xy} - \mathbf{x}_{xy} + \mu \Vert}{\gamma}}}_{\textup{gravity guidance}}, \\
    \label{eq:contact}
\end{equation}
where $c_\mathbf{v} \in \mathbb{R}$ is the contact probability of the vertex $\mathbf{v}$, $\mathbf{v}_z$ is the vertical distance of $\mathbf{v}$ to the ground, $\dot{\mathbf{v}}$ is the linear velocity of $\mathbf{v}$, $\mathbf{v}_{xy}$ is the projection of $\mathbf{v}$ on the ground, $\mathbf{x}_{xy}$ is the CoG, $\sigma$ is the sigmoidal function, and $\psi_d~,\psi_v$ are parameters to control the sigmoid slope.
The distance and velocity terms are designed similarly to prior work~\cite{brubaker09_estimcd, zell17_iccvw, zhang24_physpt}, which returns higher probability for vertices that are close to the ground and have small velocity.
The scaling of $4$ ensures the maximum probability has the value of $1$ when its vertical distance and velocity are zeros.

\begin{figure}[t]
    \centering
    \includegraphics[width=0.4\linewidth]{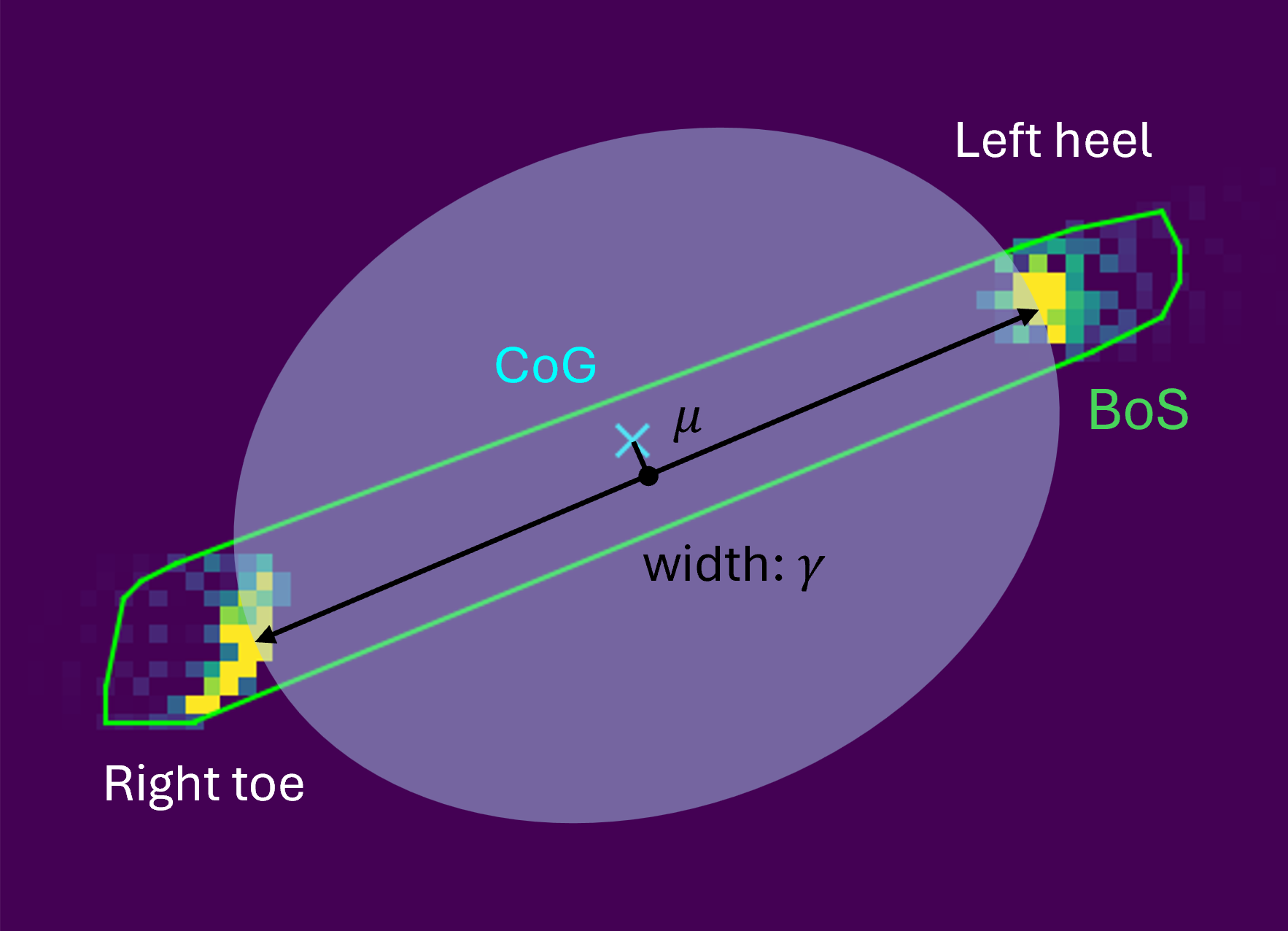}
    \caption{
    The parameters $\mu, \gamma$ (\cref{subsec:contact}) of the gravity guidance term predicted from the ConvNet.
    The offset $\mu$ compensates for the approximation error in CoM estimation.
    The $\gamma$ scales the width of the exponential function according to the base of support (BoS), the convex hull encapsulates all contact points.
    }
    \label{fig:gravity_influence}
\end{figure}

To maintain stability, the human body always tries to keep the CoG as close as possible to the CoP~\cite{winter95_balance, huec11_eqbody}, which is the average position of all body vertices weighted by their contact probabilities (Eq.~\ref{eq:cop}).
To account for this phenomenon, we propose a novel \textit{gravity guidance} term that scales the contact probability predictions accordingly to the Euclidean distance between the ground projection of every vertex $\mathbf{v}_{xy}$ and the CoG $\mathbf{x}_{xy}$.
We additionally estimate two parameters $\mu$ and $\gamma$ for the exponential function using the ConvNet.
Since each person has their own mass distribution and cannot be effectively measured, the $\mu$ compensates for the CoM approximation error from the common mass distribution from~\cite{delava96_mass}.
The \enquote{width} $\gamma$ plays an important role in scaling the distance calculation correspondingly to the base of support (BoS), which is the convex hull of all contact points on the ground. 
An example is shown in \cref{fig:gravity_influence}, during the double-support phase of a walking motion, by manipulating the $\gamma$, the exponential function widens to consider the contacts of two far-away foot contacts.

\subsection{Contact dynamics}
\label{subsec:dynamics}

Under the assumption that Newtonian mechanics is strictly enforced, all exterior forces, except gravity, applied to the body must sum up to $\mathbf{F}$, thus the sum of contact probabilities of all vertices is $1$.
However, the reaction forces in the horizontal axes $x$ and $y$ contain negative directions, which often cancel out the positive forces to maintain stability for the human body.
These cancellations cause ambiguity in the exterior force $\mathbf{F}$ computed from the CoM trajectory.
Additionally, we learn a compensation $\mathbf{F}_{\mathbf{v}}$ as additional output of the ConvNet to address this problem.
The force at each vertex $\rho_\mathbf{v} \in \mathbb{R}^{3}$ is computed as:
\begin{equation}
    \rho_\mathbf{v} = \frac{c_\mathbf{v}\exp{(c_\mathbf{v})}}{\sum^{6890}_{i=1}c_i\exp{(c_{i})}}~(\mathbf{F} + \mathbf{F}_{\mathbf{v}}),
    \label{eq:pressure}
\end{equation}
where the softmax function maps all estimated contacts to a probability distribution that sum up to $1$.
The estimated full-body contact $\mathcal{P} \in \mathbb{R}^{6890}$ is flexible towards different tasks.
We demonstrate two common types of contact dynamics that are preferred for further biomechanics research.

\textbf{Ground reaction force} (GRF) is commonly measured for two feet using force plates~\cite{han23_groundlink}.
We leverage the force plate data as supervisions denoted as $\Tilde{\lambda}_{\textup{Left}}$ and $\Tilde{\lambda}_{\textup{Right}}$.
We obtain our GRF predictions by summing all force values of the respective foot vertices, demonstrated as:
\begin{equation}
    \lambda_{\textup{Left}} = \sum^{\mathbf{V}^{\textup{Left}}}_{j} \rho_\mathbf{j}~, \quad \lambda_{\textup{Right}} = \sum^{\mathbf{V}^{\textup{Right}}}_{k} \rho_\mathbf{k}~,
    \label{eq:grf}
\end{equation}
where $\lambda_{\textup{Left}}$ and $\lambda_{\textup{Right}}$ are our predictions, $\mathbf{V}^{\textup{Left}}$ and $\mathbf{V}^{\textup{Right}}$ are the number of vertices belonging to the left and right foot, respectively.
The vertices are collected based on the segmentations from~\cite{romero17_smplh}.
The objective function $\mathcal{L}_{\textup{GRF}}$ is the mean absolute distance between $\lambda$ and $\Tilde{\lambda}$, presented as:
\begin{equation}
    \mathcal{L}_{\textup{GRF}} = \frac{1}{T} \sum^T_{t=1}\left(|\lambda_{\textup{Left}} - \Tilde{\lambda}_{\textup{Left}}| + |\lambda_{\textup{Right}} - \Tilde{\lambda}_{\textup{Right}}|\right).
\end{equation}

\textbf{A pressure map} is an another common type of contact dynamics measured via pressure mats~\cite{tripathi23_ipman}.
Unlike force plates, pressure mats only provide the force magnitude applied to the surface without direction.
We use the data from the pressure mats, which is represented as a gray-scale image $\Tilde{\mathbf{I}} \in \mathbb{R}^{n \times m}$, as supervision.
We obtain the pressure map prediction by projecting all the body contacts to the fixed area of size $n \times m$ on the ground as: $\mathbf{I} = \mathbf{A} \times \mathcal{P}$, where the projection matrix $\mathbf{A} \in \mathbb{R}^{nm\times6890}$ is constructed by activating the vertex entries with the same 2D ground-projection locations.
For each pixel in $\mathbf{I}$, the ground pressure is the sum of all vertex forces that have the same pixel coordinates, projected on the pressure mat area.
Furthermore, pressure maps also provide the CoP measurement $\Tilde{\textup{CoP}}$, and we obtain our CoP prediction via the weighted average vertex positions:
\begin{equation}
    \text{CoP} = \frac{\sum^{6890}_i\mathbf{v}_i~\Vert\rho_i\Vert}{\sum^{6890}_i\Vert\rho_i\Vert}.
    \label{eq:cop}
\end{equation}

The objective for training with pressure mat data $\mathcal{L}_{\textup{mat}}$, consisting of both ground pressure map and CoP prediction, is computed as:
\begin{equation}
    \mathcal{L}_{\textup{Mat}} = \frac{1}{T} \sum^T_{t=1} \left( (\mathbf{I}_t - \Tilde{\mathbf{I}}_t)^2 + (\textup{CoP} - \Tilde{\textup{CoP}})^2 \right)~.
\end{equation}

\subsection{Training objectives}
\label{subsec:objectives}

Overall, the full objective function for training the ConvNet is shown as:
\begin{equation}
    \mathcal{L} = \mathcal{L}_{\textup{CoM}} + \mathcal{L}_{\textup{GRF}} + \mathcal{L}_{\textup{Mat}} + \Lambda \mathcal{L}_{\textup{reg}},
    \label{eq:loss}
\end{equation}
where the regularization $\mathcal{L}_{\textup{reg}} = \log(\gamma)^2 $ is to penalize $\gamma$ collapsing to zero, preventing the model to project all pressures into one vertex; and $\Lambda$ is the weighting factor of $\mathcal{L}_{\textup{reg}}$ towards the total objective.
Besides the compulsory CoM reconstruction loss $\mathcal{L}_{\textup{CoM}}$, $\mathcal{L}_{\textup{GRF}}$ and $\mathcal{L}_{\textup{Mat}}$ are activated based on the availability of the respective sensory measurements.

\section{Experiments}
\label{sec:experiments}

\subsection{Datasets}
\label{subsec:datasets}

We conduct the experiments on two datasets.
The first one is GroundLink~\cite{han23_groundlink}, which contains motion-capture recordings of a wide variety of daily and athletic motions performed by $7$ actors.
Unlike most datasets included in the AddBiomechanics collection~\cite{werling2024addbiomechanics}, which predominantly consist of walking trials, GroundLink provides substantial motion diversity, making it a more challenging and representative benchmark for GRF and CoP estimation.
All motions are accompanied by the corresponding force-plate measurements of GRF and CoP.
Based on the reported results in~\cite{han23_groundlink}, we split the data subject-wise, with $s1, s2, s3, s4, s6$ for training and $s5, s7$ for testing.
The second dataset is the MOYO~\cite{tripathi23_ipman}, which contains $200$ complex yoga motions performed by one actor, with corresponding pressure mat and CoP measurements.
We follow the data split provided in the dataset~\cite{tripathi23_ipman} for evaluation.
We additionally evaluate GraCE on the Fit3D dataset~\cite{fieraru2021_fit3d} which contains motion capture data of sport movements.
Since Fit3D provides no force plate measurements, only qualitative results are presented.
All three datasets provided SMPL-X~\cite{pavlakos19_smplx} pose parameters fitted to the motion capture markers via MoSh~\cite{loper14_mosh}.
Since facial expressions are not regarded for contact dynamics estimation, we only use the body and hand poses to obtain the input SMPL+H motions.

\subsection{Implementation details}
\label{subsec:details}

For a fair comparison, related work~\cite{han23_groundlink, liu25_imdy, brubaker09_estimcd, zhang24_physpt, tripathi23_ipman} is re-implemented and evaluated with the same experiment configuration.
The learning component of GraCE is the ConvNet, which has the same architecture as the GLinkNet~\cite{han23_groundlink},  consisting of four 1D convolutional layers of kernel size $7$ and a hidden dimension of 128, followed by three fully connected layers of dimension 256.
Instead of outputting the GRF and CoP directly as~\cite{han23_groundlink}, we predict the control parameters $\kappa_P$, $\kappa_D$, and $b$ for the CoM reconstruction; and $\mu$, $\gamma$ for the scaling factors of the proposed gravity-guided contact model.
The sigmoid slopes $\psi_d=60~, \psi_v=10$ are selected similarly to previous work~\cite{brubaker09_estimcd, zhang24_physpt} for a fair comparison.
The input to all models is the SMPL parameters ($\mathbf{x}, \mathbf{y}, \theta, \beta$) concatenated to one vector.

All motions from the GroundLink and MOYO datasets are down-sampled to $50$Hz.
The models are equally trained for $25$ epochs with a batch size of $8$ and a learning rate of $10^{-4}$.
Each neural linear layer is followed by an Exponential Linear Unit (ELU) activation function.
During the training of GraCE, the regularization loss $\mathcal{L}_{\text{reg}}$ is scaled by $\Lambda=0.01$.
The time step $\Delta t=0.02$ corresponds to the down-sampled capture rate of $50$Hz.
All experiments are conducted on five different random seeds $0-4$ and we report the mean and standard deviation.
Following the standard protocol in prior literature~\cite{han23_groundlink,liu25_imdy,brubaker09_estimcd,zell17_iccvw}, all reported GRF predictions are normalized by the person's body weight BW (\unit{\kilo\gram}$\cdot$\unit{\metre}/\unit{\second}$^2$).

\subsection{Metrics}
\label{subsec:metrics}

We first evaluate the GRF and CoP predictions on the force-plate measurements in the GroundLink dataset.
We compute the ground force error for left and right foot contacts (GFE\textsubscript{L} and GFE\textsubscript{R}) as the average squared error between the GRF predictions and force-plate data (in N/BW) in the respective foot.
Since the dataset contains mostly motions with standing-up pose, the GRF in z-axis is the most interesting and reported separately as GFE\textsubscript{z}.
The error in CoP prediction is measured by the center of pressure error (CPE), which is the average L2 distance (in \unit{\milli\metre}) between the predicted CoP and the $\Tilde{\text{CoP}}$ measured from force plates.
The CPE in left and right feet are reported as CPE\textsubscript{L}, CPE\textsubscript{R}, and the average is $\overline{\text{CPE}}$.
For evaluation on MOYO, we compute the average CPE between the CoP of full-body estimation from GraCE and the pressure mat measurements~\cite{tripathi23_ipman}.

\subsection{Results}
\label{subsec:results}

\textbf{On GroundLink}.
We present the results of GraCE on the GroundLink dataset in~\cref{tab:results_glink}.
The related work consists of two categories: 1) data-driven methods with GLinkNet~\cite{han23_groundlink} with a CNN, and ImDyS~\cite{liu25_imdy} with a transformer; and 2) explicit contacts from SMPL with distance-based models EstimCD~\cite{brubaker09_estimcd}, IPMAN~\cite{tripathi23_ipman}, and a distance-velocity-based model PhysPT~\cite{zhang24_physpt}.
GraCE achieves state-of-the-art results and significantly outperforms the closest method IPMAN by $40.01\%$ in GFE\textsubscript{z}, $50.27\%$ in GFE\textsubscript{L}, and $32.93\%$ GFE\textsubscript{R}.
Compared to a model that also considers vertex velocity in the modeling of contact, PhysPT, GraCE reduces the GFE\textsubscript{z} by $53.01\%$ due to the additional gravity-guided estimation in~\cref{eq:contact}.
Regarding the quality of CoP estimation, the data-driven methods, GLinkNet and ImDyS, are clearly inferior to explicit methods with over $50\%$ higher CPE.
The better estimation of the explicit contact models is due to the SMPL body constraint, where the prediction of CoP from the feet cannot go beyond the body limitation.
Compared to other models, we achieve the lowest average CPE of $56.19\unit{\milli\metre}$, $11.45\%$ lower than the closest EstimCD with $63.46\unit{\milli\metre}$.

\begin{table}[t]
    \centering
    \caption{
    Quantitative results on the GroundLink dataset~\cite{han23_groundlink}.
    All methods are trained and evaluated on the same training configuration.
    Lower error means better performance.
    Best evaluation results on respective metrics are in \textbf{bold}.
    }
    \begin{tabularx}{\linewidth}{l@{\extracolsep{\fill}}|ccc|ccc}
        \toprule
        Method & GFE\textsubscript{z} & GFE\textsubscript{L} & GFE\textsubscript{R} & CPE\textsubscript{L} & CPE\textsubscript{R} & $\overline{\text{CPE}}$ \\
        \midrule
        GLinkNet~\cite{han23_groundlink}  & 5.13 & 1.72 & 1.96 & 134.52 & 141.39 & 137.95 \\
        ImDyS~\cite{liu25_imdy}           & 5.46 & 2.07 & 1.79 & 69.22 & 133.20 & 101.21 \\
        PhysPT~\cite{zhang24_physpt}      & 5.81 & 2.15 & 2.13 & 55.13 & 76.02 & 65.57 \\
        EstimCD~\cite{brubaker09_estimcd} & 5.34 & 1.85 & 2.13 & \textbf{53.70} & 73.22 & 63.46 \\
        IPMAN~\cite{tripathi23_ipman}     & 4.55 & 1.81 & 1.67 & 64.99 & 65.21 & 65.10 \\
        GraCE~(ours) & \textbf{2.73}\tiny{$\pm$0.14} & \textbf{0.9}\tiny{$\pm$0.07} & \textbf{1.12}\tiny{$\pm$0.02} & 54.17\tiny{$\pm$0.61} & \textbf{58.21}\tiny{$\pm$1.38} & \textbf{56.19} \\
        \bottomrule
    \end{tabularx}
    \label{tab:results_glink}
\end{table}

\begin{figure}[t!]
    \centering
    \begin{subfigure}[b]{0.48\textwidth}
        \centering
        \includegraphics[width=\linewidth]{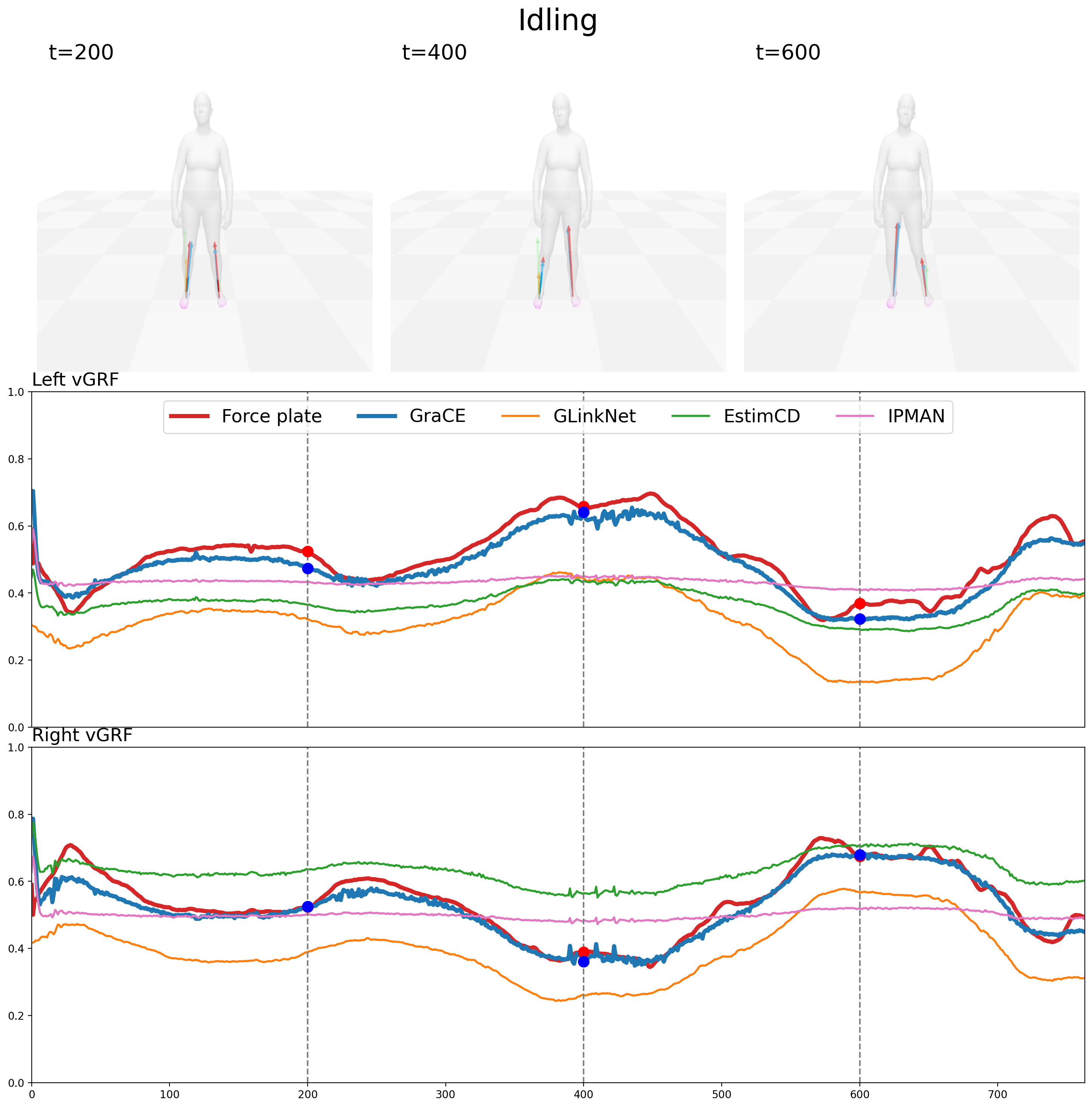}
    \end{subfigure}
    ~ 
    \begin{subfigure}[b]{0.48\textwidth}
        \centering
        \includegraphics[width=\linewidth]{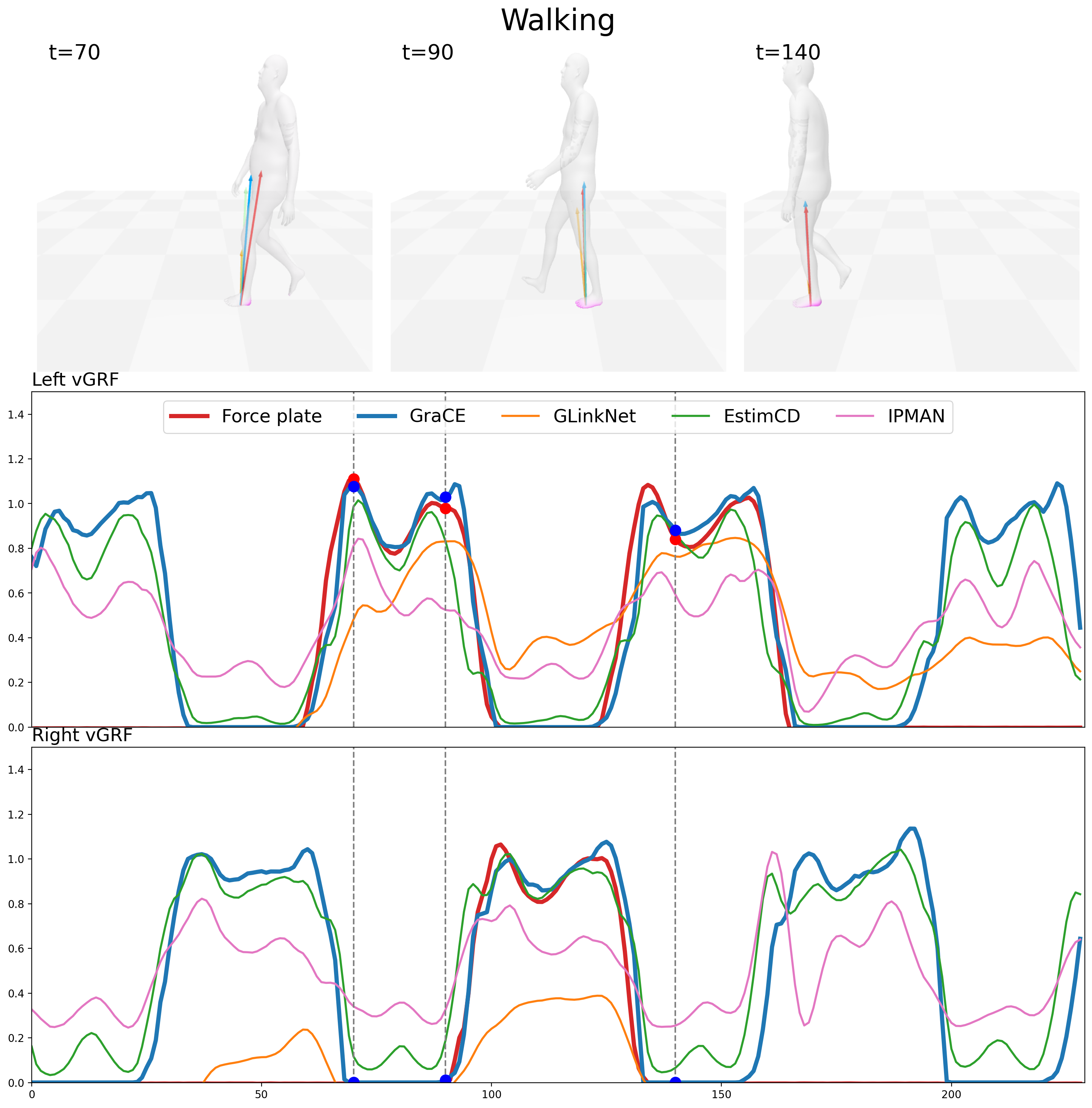}
    \end{subfigure}
    \\
    \begin{subfigure}[b]{0.48\textwidth}
        \centering
        \includegraphics[width=\linewidth]{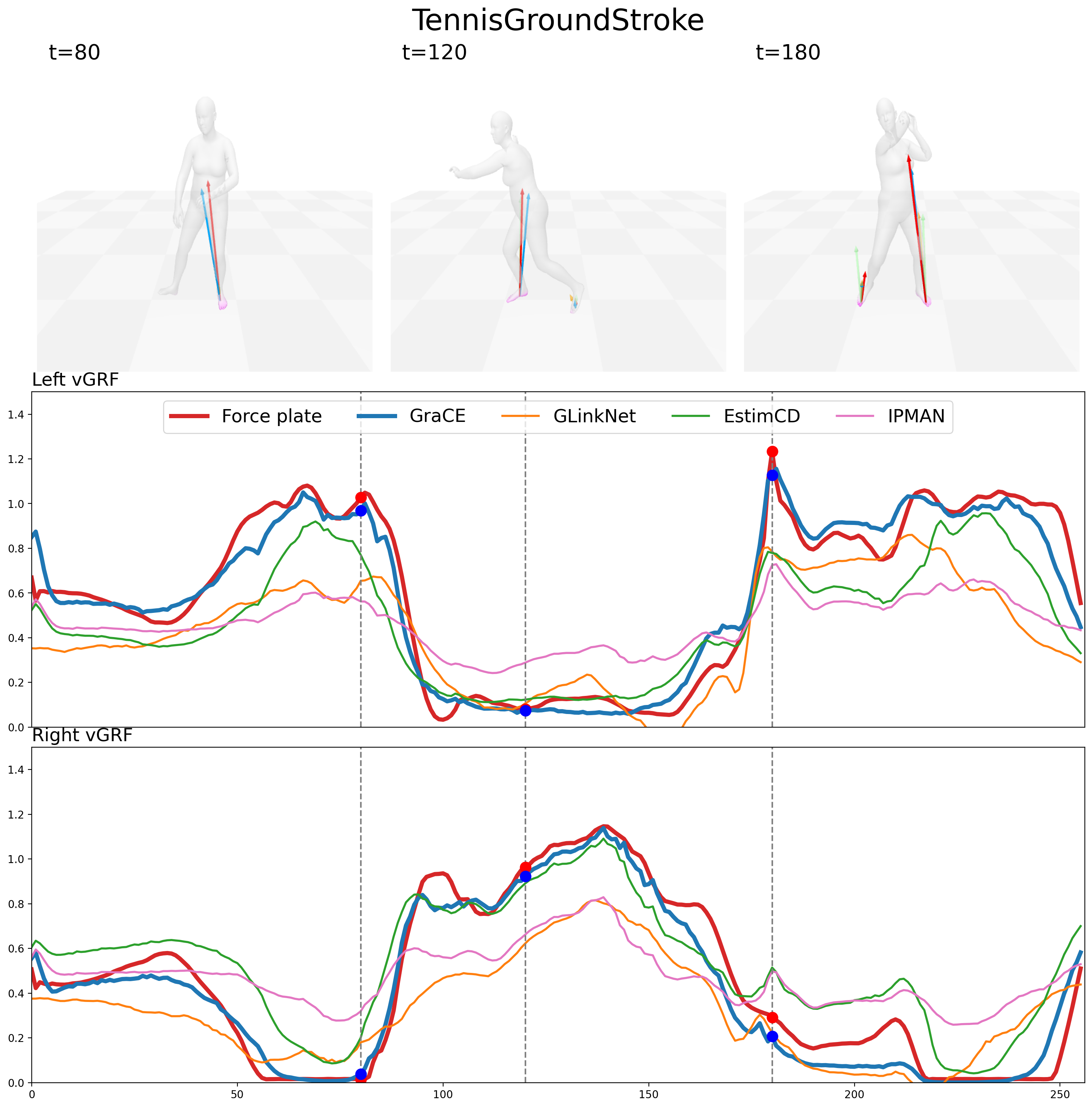}
    \end{subfigure}
    ~ 
    \begin{subfigure}[b]{0.48\textwidth}
        \centering
        \includegraphics[width=\linewidth]{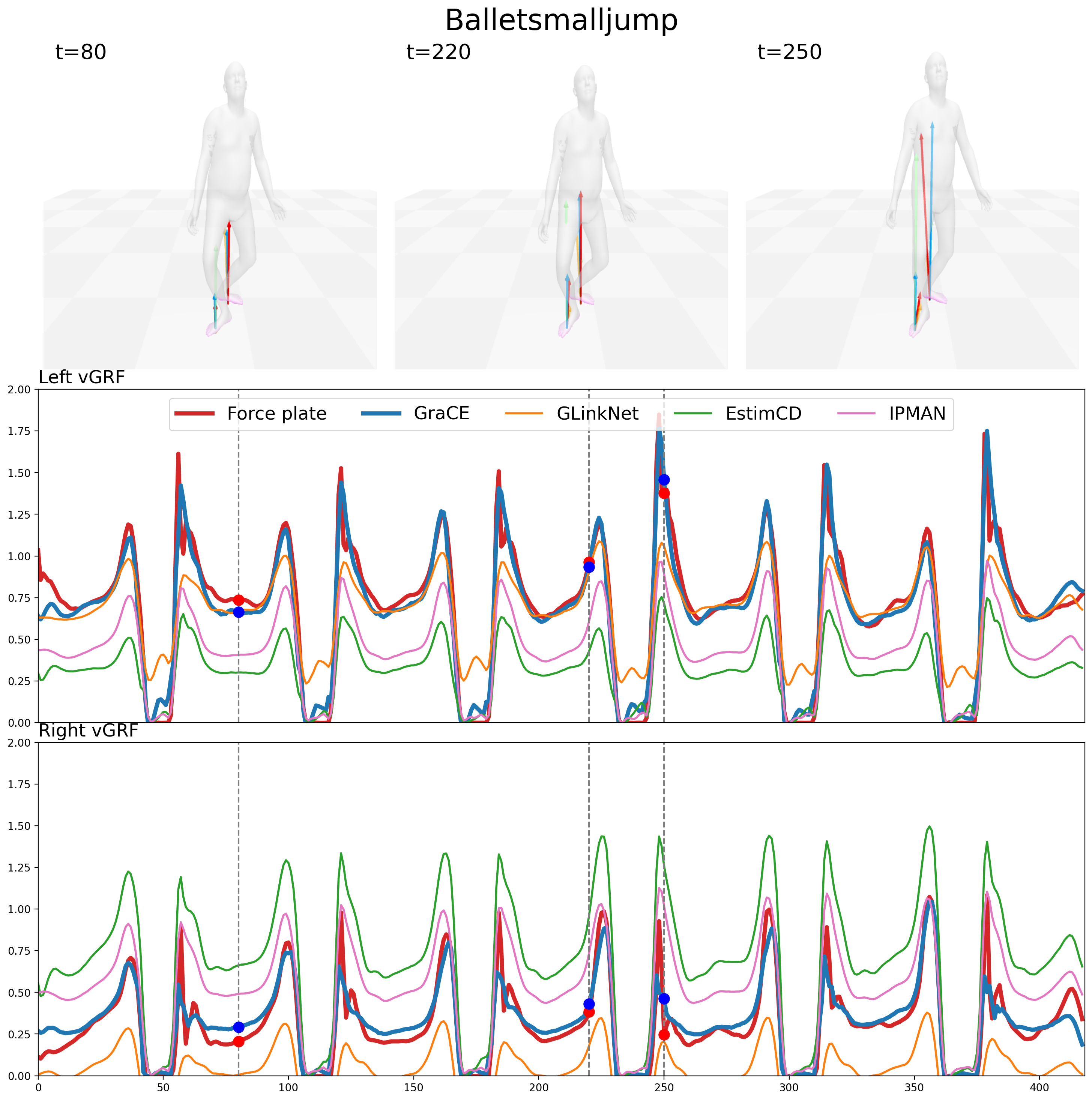}
    \end{subfigure}
    \caption{
    Qualitative results of GraCE, along with related work on GroundLink dataset~\cite{han23_groundlink}.
    We plot the vertical GRF estimations of the left and right feet below each example motion.
    In each example, we visualize the human pose and the GRF vector arrows with colors corresponding to the graph legend.
    Unlike previous work \cite{han23_groundlink, liu25_imdy}, we show the raw estimations from the models without any temporal filtering.
    All contact dynamics predictions are normalized by the person's weight.
    }
    \vspace{-10pt}
    \label{fig:qualitative_glink}
\end{figure}

Qualitative results of GraCE in comparison with related methods are shown in~\cref{fig:qualitative_glink}.
The benefit of using gravity-guided estimation in GraCE is shown in the example of an idling motion on the top left.
In this motion, the human body is swinging left and right while the position (and velocity) of the two feet in contact do not change much, causing distance-based methods, such as EstimCD (green) or IPMAN (pink), fail to capture the correct dynamics and predict constant GRF.
In contrast, our GraCE is guided by the CoG, which takes into account the swinging movement of the body, correctly predicts the GRF (blue) w.r.t the force plate signals (red).
In the second example of a walking motion in the top right of~\cref{fig:qualitative_glink}, GraCE also outperforms other methods in accuracy of contact dynamics.
Moreover, due to sensor limitation, force plate data is only available in a fixed working area, \ie the red graph only has values from time step 60 to 160 in the walking example.
GraCE effectively addresses this problem by providing high quality contact dynamics estimations directly from the motion capture data, which has a wider range of applications.
On more complex motions such as tennis stroke (\cref{fig:qualitative_glink}, bottom left) and ballet small jump (bottom right), with the guidance from the CoG, GraCE can correctly capture the peak force signals when the person applies high pressure to the ground.

\begin{figure}[t!]
    \centering
    \begin{subfigure}[b]{0.48\textwidth}
        \centering
        \includegraphics[width=\linewidth]{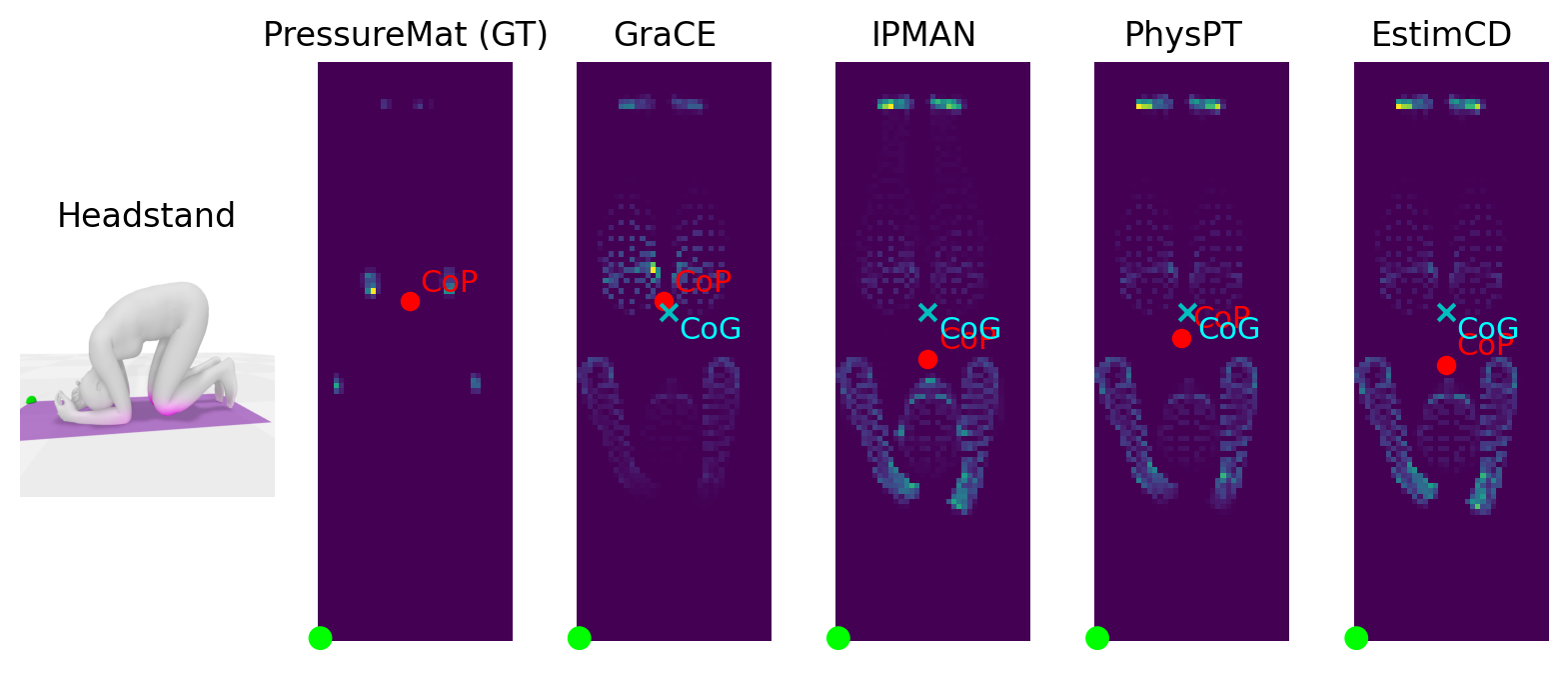}
    \end{subfigure}
    ~
    \centering
    \begin{subfigure}[b]{0.48\textwidth}
        \centering
        \includegraphics[width=\linewidth]{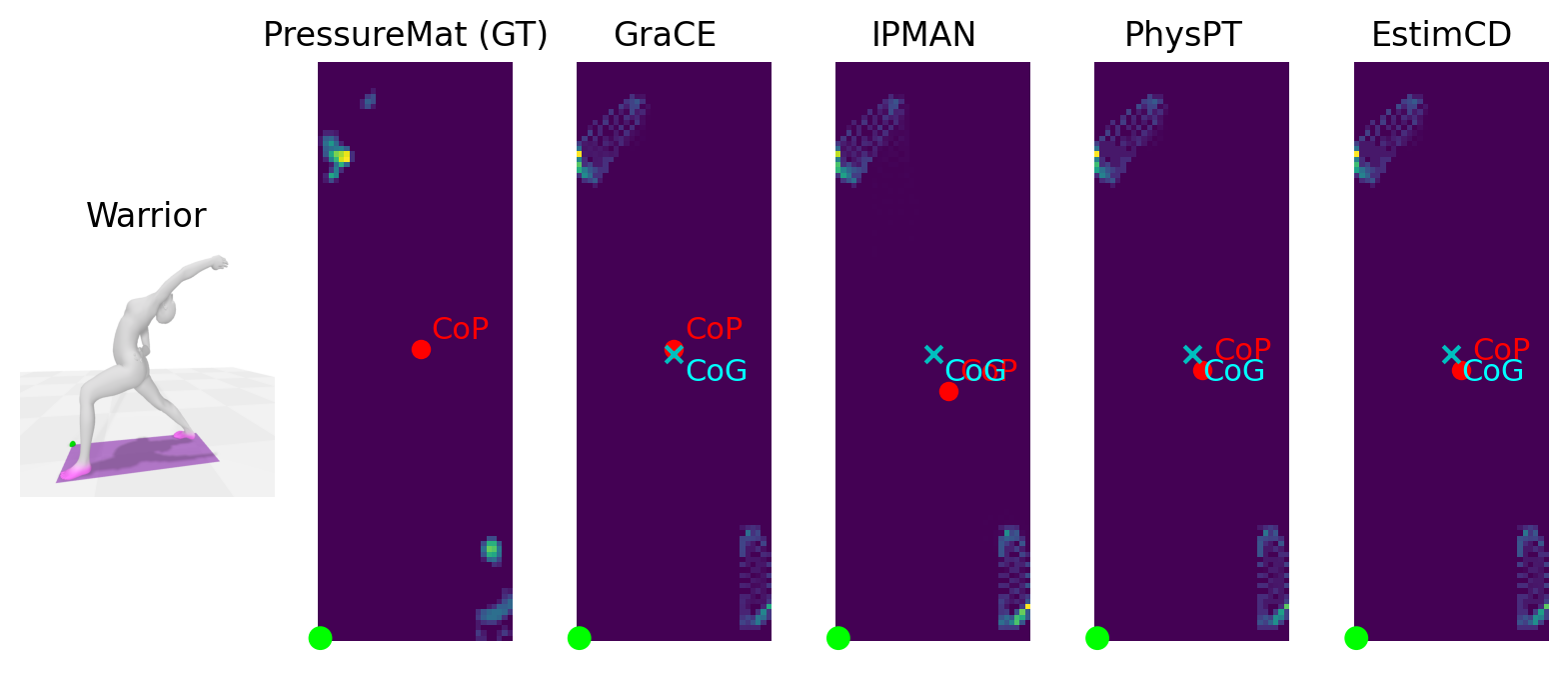}
    \end{subfigure}
    \\
    \begin{subfigure}[b]{0.48\textwidth}
        \centering
        \includegraphics[width=\linewidth]{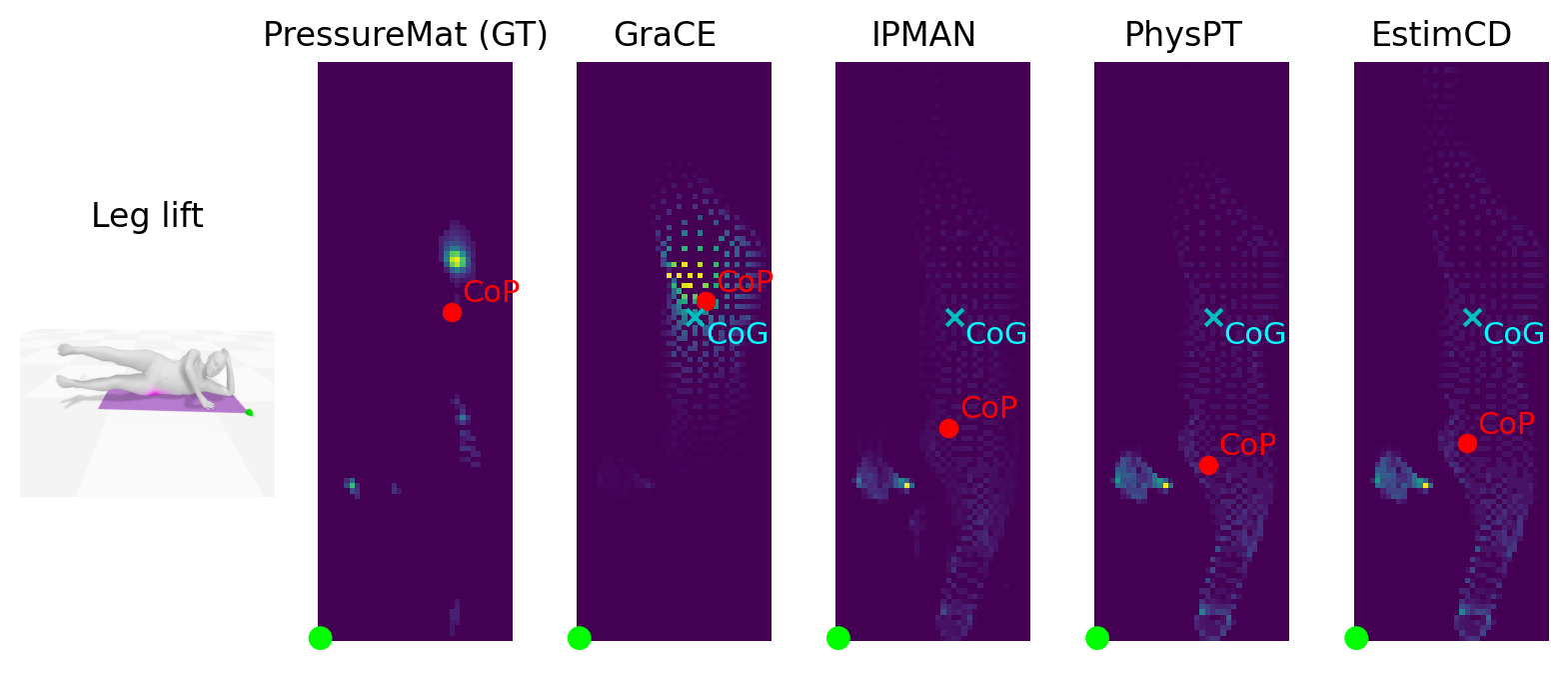}
    \end{subfigure}
    ~
    \centering
    \begin{subfigure}[b]{0.48\textwidth}
        \centering
        \includegraphics[width=\linewidth]{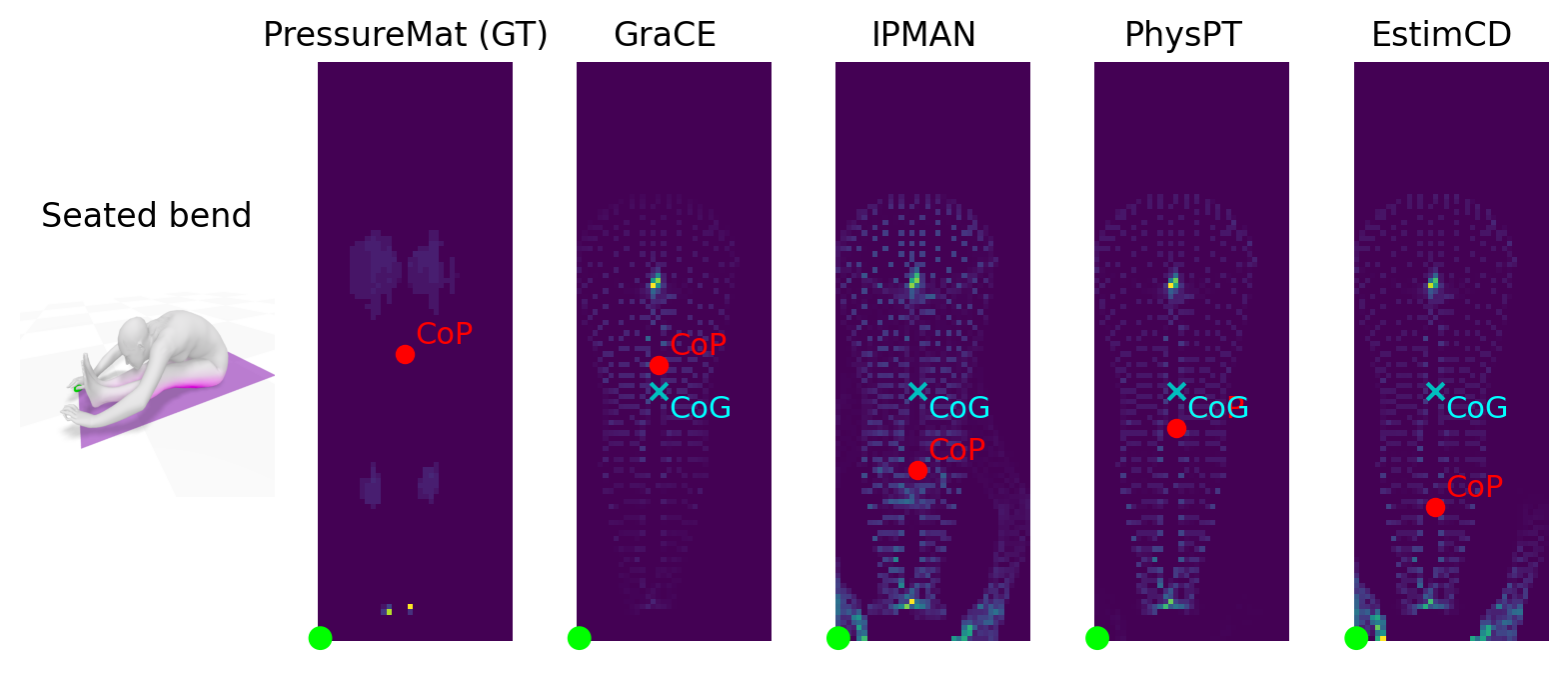}
    \end{subfigure}
    \caption{
    Qualitative results on the MOYO dataset~\cite{tripathi23_ipman}.
    One frame of the motion is shown on the left of each sample with the contact pressure highlighted in {\color{Magenta}{violet}}.
    The pressure maps of all contact models in comparison, produced by projecting all vertex pressures to the ground plane, are demonstrated from left to right: GT from pressure mat sensors, GraCE, IPMAN~\cite{tripathi23_ipman}, PhysPT~\cite{zhang24_physpt} and EstimCD~\cite{brubaker09_estimcd}.
    The estimated CoP of each map is denoted in {\color{red}{red}}, the origin of the pressure mat is denoted in {\color{Green}{green}}.
    The CoG ({\color{cyan}{cyan}}) is shown in the pressure maps produced by GraCE to demonstrate the high correlation between CoP and CoG in maintaining good estimations.
    }
    \label{fig:qualitative_moyo}
\end{figure}

\textbf{On MOYO}.
We present the quantitative CoP estimation results on the MOYO dataset in~\cref{tab:results_moyo}.
On average, GraCE achieves lower CoP estimation error than other related contact models, with a decrease of $9.1\%$ compared to the closest method IPMAN.
In~\cref{fig:qualitative_moyo}, we show four examples where GraCE excels in the estimation of CoP compared to distance-based approaches.
In all four cases, the estimations of the CoP from GraCE closely match the CoP measured by the pressure mat sensor.
In the headstand pose (top left), the majority of the pressures are propagated to the knees even though the vertices in the head and arm areas are detected to be at a lower vertical height, and this is reflected via the CoG of the human body.
Based on the CoG, GraCE produces a more accurate estimation than the distance-based approaches.
Similar with other examples, \ie, leg lift (bottom left) or seated bend (bottom right) in~\cref{fig:qualitative_moyo}, where distance-based estimations are biased towards the vertically lower body parts (the hands), not where the pressure is being applied (the hip).

\begin{table}[t]
    \centering
    \setlength{\tabcolsep}{8pt}
    \caption{
    Quantitative results of CPE w.r.t to pressure mat measurements on the MOYO~\cite{tripathi23_ipman}.
    The best average result is shown in \textbf{bold}.
    The CoP errors are in \unit{\milli\metre}.
    }
    \begin{tabular}{l|rrrr}
        \toprule
        Yoga pose & EstimCD & PhysPT & IPMAN & GraCE \\
        \midrule
        Leg lift        & 146.75 & 160.09 & 134.35 & 122.20\tiny{$\pm$8.73} \\
        Bound angle     & 76.95  & 78.87  & 77.42  & 75.15\tiny{$\pm$1.13}  \\
        Seated bend     & 161.80 & 138.96 & 135.69 & 88.18\tiny{$\pm$3.18}  \\
        Plank           & 86.40  & 87.12  & 65.30  & 71.03\tiny{$\pm$3.96}  \\
        Headstand       & 89.82  & 85.58  & 83.90  & 64.75\tiny{$\pm$3.78}  \\
        Bow pose        & 110.92 & 109.61 & 120.74 & 103.40\tiny{$\pm$9.88} \\
        Warrior II      & 64.69  & 63.45  & 55.25  & 62.08\tiny{$\pm$0.73}  \\
        Warrior III     & 50.87  & 47.17  & 50.93  & 46.18\tiny{$\pm$0.06}  \\
        Reverse warrior & 64.26  & 60.47  & 66.48  & 59.58\tiny{$\pm$2.14}  \\
        Diamond pose    & 117.27 & 119.79 & 112.78 & 110.02\tiny{$\pm$2.95} \\
        Tree pose       & 29.79  & 29.37  & 28.17  & 29.42\tiny{$\pm$0.02}  \\
        \midrule
        Average         & 87.78 & 86.36 & 84.0 & \textbf{76.33}\tiny{$\pm$0.91} \\
        \bottomrule
    \end{tabular}
    \label{tab:results_moyo}
\end{table}

\begin{figure}[t]
    \centering
    \includegraphics[width=0.85\linewidth]{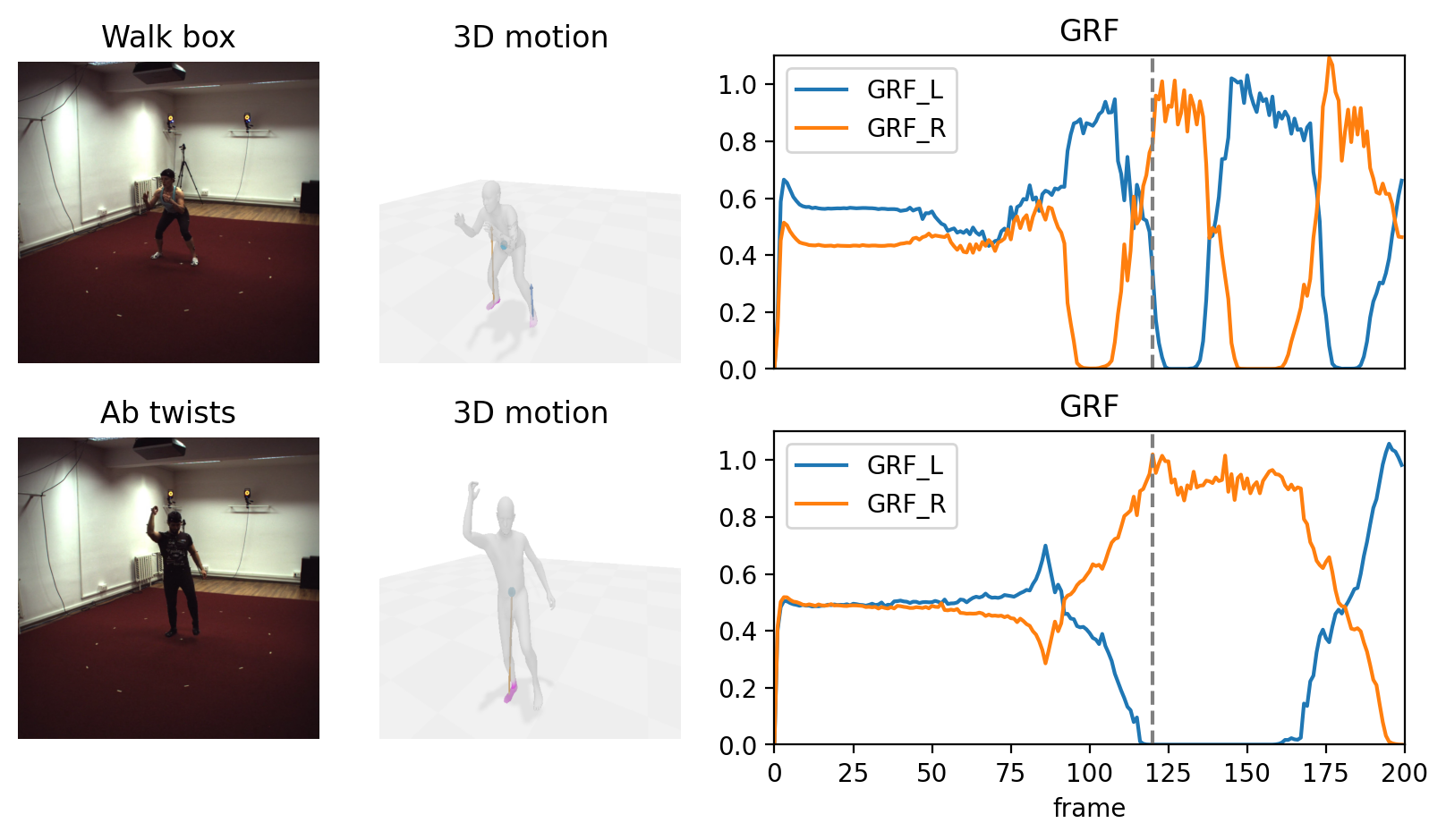}
    \caption{
    Qualitative results on the Fit3D~\cite{fieraru2021_fit3d}.
    The footage is shown on the top with their corresponding motion captures below.
    The contacts are denoted with purple, the ground reaction forces are represented with arrows and the CoM is the blue sphere.
    }
    \label{fig:qualitative_fit3d}
    \vspace{-15pt}
\end{figure}

\textbf{On Fit3D}.
The qualitative results on the Fit3D sport motions in~\cref{fig:qualitative_fit3d} demonstrates the plausibility of contact dynamics of GraCE.
The magnitude of vertex contact pressure, together with the estimated ground reactions are correctly computed based on the position of the CoM.
For instance, in the walk box motion, even though the two feet are at the same vertical level, the estimated force in the right foot is higher due to the hip position slightly leaning right, which is captured by the CoM.

\subsection{Ablation studies}
\label{subsec:ablation}

In~\cref{tab:ablations}, we verify the configuration of GraCE by isolating out the proposed components individually and report their final performance on the GroundLink dataset.
All the performance gains (in $\%$) below is for the Full GraCE model compared to the respective isolated components.

\textbf{Motion-based exterior force}.
The usage of a PD controller for estimating the exterior force $\mathbf{F}$ from the CoM trajectory is vital for the capture of contacts dynamics.
Without the $\mathbf{F}$ and using only $\mathbf{F}_{\mathbf{v}}$, the GraCE model \textit{w/o PD algorithm} cannot produce meaningful reaction forces, leading to a large value of $37.29\text{N/BW}$ in GFE\textsubscript{z}.
The usage of Softmax is to strictly ensure that the Newton mechanics in~\cref{subsec:exterior} holds.
In the model configuration \textit{w/o Softmax}, all contact forces are not guaranteed to sum up to the total exterior force that controls the CoM movements, leading to a poor performance with also $37.26\text{N/BW}$ in GFE\textsubscript{z}.
The learnable $\mathbf{F}_{\mathbf{v}}$, demonstrated via the GraCE configuration \textit{w/o learning $\mathbf{F}_{\mathbf{v}}$}, is important to recover the ambiguous negative ground reaction applied on the horizontal axes, with $25\%$ and $16\%$ decreases in GFE\textsubscript{L} and GFE\textsubscript{R}, respectively.

\textbf{Gravity-guided estimation}.
Our proposed gravity-guided term contributes greatly to the accuracy of contact dynamics estimation compared to the baseline distance-velocity-based model (\textit{w/o CoG guidance}), with a significant $48.9\%$ decrease in GFE\textsubscript{z}, $48.5\%$ in GFE\textsubscript{L}, and $44.5\%$ in GFE\textsubscript{R}.
The parameters $\mu$ and $\gamma$ for controlling the influence of the CoG are important to adaptively: 1) correct the CoM approximation error due to ambiguous person-specific mass distribution; and 2) modify the steepness of the exponential function when encountering different widths of the BoS.
The configuration \textit{w/o learning $\mu, \gamma$} is verified by setting $\mu=0$ and $\gamma=1$ and receives a lower performance in GFE compared to the full model by $44.7\%$ in GFE\textsubscript{z}, $44.1\%$ in GFE\textsubscript{L}, and $40.7\%$ in GFE\textsubscript{R}.

\textbf{CoP estimation} quality in the left foot (CPE\textsubscript{L}) differs slightly between configurations due to the high number of simple motions in the testing set with the ``tree pose'' standing on a single left foot.
GraCE's ability to capture more complex CoP distributions is demonstrated via the right foot in CPE\textsubscript{R}, where the full model outperforms the closest GraCE configuration by $2.0\%$, and a significant $13.6\%$ compared to the closest related work IPMAN in~\cref{tab:results_glink}.

\begin{table}[t]
    \centering
    \caption{
    Ablation studies on different model configurations.
    All proposed components of GraCE are isolated individually to study their contribution towards the full model.
    Best results are highlighted with \textbf{bold}, and second best are \underline{underlined}.
    }
    \begin{tabularx}{\linewidth}{l@{\extracolsep{\fill}}rrrrrr}
        \toprule
        GraCE config. & GFE\textsubscript{z} & GFE\textsubscript{L} & GFE\textsubscript{R} & CPE\textsubscript{L} & CPE\textsubscript{R} & $\overline{\text{CPE}}$ \\
        \midrule
        ~\textit{w/o} PD algorithm           & 37.29\tiny{$\pm$0.01} & 17.01\tiny{$\pm$0.02} & 8.08\tiny{$\pm$0.02} & 59.66\tiny{$\pm$2.21} & \textbf{56.01}\tiny{$\pm$1.79} & 57.83 \\
        ~\textit{w/o} Softmax                & 37.26\tiny{$\pm$0.01} & 17.12\tiny{$\pm$0.01} & 8.19\tiny{$\pm$0.02} & 67.25\tiny{$\pm$0.01} & 61.68\tiny{$\pm$0.01} & 64.46 \\
        ~\textit{w/o} learning $\mathbf{F}_{v}$       & \underline{3.23}\tiny{$\pm$0.66} & \underline{1.22}\tiny{$\pm$0.19} & \underline{1.33}\tiny{$\pm$0.25} & 55.46\tiny{$\pm$2.37} & 59.18\tiny{$\pm$1.99} & \underline{57.32} \\
        ~\textit{w/o} CoG guidance           & 5.35\tiny{$\pm$0.05} & 1.77\tiny{$\pm$0.01} & 2.02\tiny{$\pm$0.03} & \underline{53.49}\tiny{$\pm$0.01} & 73.71\tiny{$\pm$0.01} & 63.60 \\
        ~\textit{w/o} learning $\mu, \gamma$ & 4.94\tiny{$\pm$0.03} & 1.63\tiny{$\pm$0.01} & 1.89\tiny{$\pm$0.03} & \textbf{53.02}\tiny{$\pm$0.01} & 72.19\tiny{$\pm$0.01} & 62.61 \\
        ~\textit{w/o} learning $\mu$         & 4.02\tiny{$\pm$0.09} & 1.29\tiny{$\pm$0.02} & 1.61\tiny{$\pm$0.05} & 53.50\tiny{$\pm$0.01} & 71.69\tiny{$\pm$0.01} & 62.59 \\
        ~\textit{w/o} learning $\gamma$      & 4.98\tiny{$\pm$0.05} & 1.63\tiny{$\pm$0.02} & 1.91\tiny{$\pm$0.03} & 53.02\tiny{$\pm$0.01} & 72.19\tiny{$\pm$0.01} & 62.61 \\
        Full model & \textbf{2.73}\tiny{$\pm$0.14} & \textbf{0.91}\tiny{$\pm$0.07} & \textbf{1.12}\tiny{$\pm$0.02} & 54.17\tiny{$\pm$0.61} & \underline{58.21}\tiny{$\pm$1.38} & \textbf{56.19} \\
        \bottomrule
    \end{tabularx}
    \label{tab:ablations}
\end{table}

\section{Conclusion}
\label{sec:conclusion}
In this paper, we present a novel gravity-guided model, GraCE, for contact dynamics estimation directly from motion captures.
By integrating our novel gravity guidance, we are able to significantly improve the estimation accuracy of ground reaction dynamics, which was not possible in previous work that only concern the vertical distance and velocity to compute contact probability.
GraCE demonstrates a state-of-the-art performance in accuracy of ground reaction force and pressure estimations on the two datasets, GroundLink and MOYO.

\paragraph{Limitation and future work.}
The current SMPL models used in GraCE do not deform under pressure, leading to a high reliance of the dynamics model on the precision of the motion captures.
Deformation that directly results from the interactions between body vertices with the ground will be investigated in the future work.
Additionally, while the current design of GraCE can effectively recover the reaction forces from quasi-static human motions, it cannot fully capture the forces if no motion is presented, \eg, the person intentionally applies high forces to the surface but creates no motion.
Addressing this ambiguous problem from visual cues alone is challenging, and more sophisticated modelings of the environmental contexts, such as the contact surface density, the precise mass of the person, and muscle activities, could benefit the estimations of GraCE.

\section*{Acknowledgment and disclosure of funding} 
This research is supported by the Wallenberg Artificial Intelligence, Autonomous Systems and Software Program (WASP), funded by Knut and Alice Wallenberg Foundation.
The computational resources were provided by the National Academic Infrastructure for Supercomputing in Sweden (NAISS) at C3SE, and by the Berzelius resource, provided by the Knut and Alice Wallenberg Foundation at the National Supercomputer Centre.

%
%
\clearpage
\bibliographystyle{splncs04}
\bibliography{references}

\newpage
\clearpage
\maketitlesupplementary

This supplementary document provides additional information about the proposed method GraCE.
In Sec.~\ref{supplsec:additional}, we provide the descriptions about the additional qualitative results provided in the~\textit{index.html} file.
Sec.~\ref{supplsec:com} contains the calculation of the center of mass (CoM) from a SMPL+H mesh.
Sec.~\ref{supplsec:convnet} provides more details about the ConvNet and the parameter selection.

\section{Additional qualitative results}
\label{supplsec:additional}

We present the comprehensive videos in the supplementary~\textit{index.html}, demonstrating the ground contact dynamics estimations for all three datasets presented in the main paper.
Each example from the GroundLink dataset is accompanied by a ground pressure map and the reaction forces estimation from GraCE at contact points w.r.t the force plate measurements.
The full-body predictions of GraCE on the MOYO dataset, projected to the area of the pressure mat sensor, are visually compared to the sensor measurements.
We additionally provide the ground reaction force estimations on the Fit3D dataset in the last two columns, with the arrows corresponding to the magnitude and direction of the forces.

\section{CoM estimation}
\label{supplsec:com}

\begin{figure}[ht]
    \centering
    \includegraphics[width=0.8\linewidth]{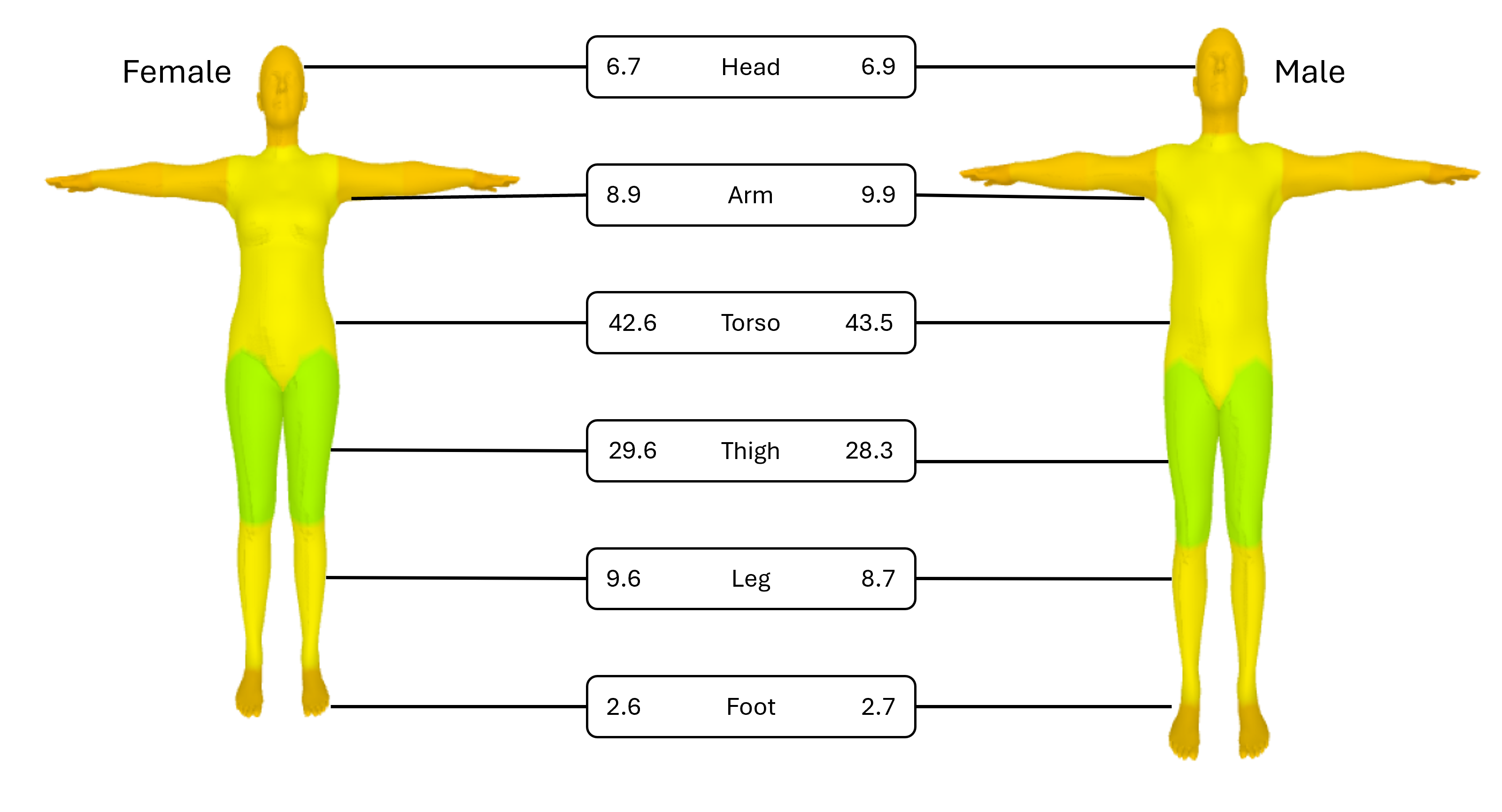}
    \caption{
    Common body mass distribution from~\cite{delava96_mass} for female and male.
    The numbers in the middle show the percentage of mass of specific body parts compared to the person's total mass normalized to one.
    The female and male bodies have slight differences between the parts, but both of them sum up to 100\%.
    }
    \label{supplfig:mass}
\end{figure}

Giving a SMPL mesh $\mathbf{V} \in \mathbb{R}^{6890\times3}$, we compute the body center of mass (CoM), $\mathbf{r}\in\mathbb{R}^{3}$, using the common mass distribution from~\cite{delava96_mass}, where specific mass percentages of body parts can be seen in Fig.~\ref{supplfig:mass}.

We first find the mass percentage of each vertex, $\omega_{\mathbf{v}}$, out of the total $6890$ vertices by dividing the mass percentage of each body part by the number of vertices that belong to that specific part defined by~\cite{romero17_smplh}.
The computation of the CoM is the average of all $6890$ vertices, weighted by the their mass percentages:
\begin{equation}
    \mathbf{r} = \sum^{6890}_{j} \omega_{\mathbf{v}_{j}}~\mathbf{v}_{j}~,
    \label{suppleeq:com}
\end{equation}
where $\mathbf{v}_{j}$ is the 3D position of one vertex and $\omega_{\mathbf{v}_{j}}$ is the corresponding mass percentage of that vertex.
The projection of the CoM to the ground, the center of gravity (CoG), is found by setting the z-axis coordinate of CoM to zero.

\section{Details about the ConvNet}
\label{supplsec:convnet}

For fair comparisons, we use the same backbone architecture as GLinkNet~\cite{han23_groundlink}, which is demonstrated in Sec.~4.2 in the main paper.
There are three output heads for our ConvNet: PD controller's parameters, learned vertex pressure, and the parameters for the gravity influence term.
The PD controller's parameters $\kappa_P, \kappa_D, b$ are respectively bounded by the ranges $[0, 80]$, $[0, 40]$ and $[0, 10]$.
This helps with stabilizing the simulation results at the beginning of training when the parameters are not fully optimized.
Similarly, the CoM compensation $\mu$ is bounded in the range of $[-0.2, 0.2]$, restricting the prediction to be within a radius of $20\unit{\centi\metre}$ from the estimation CoM from~\cref{suppleeq:com}.
The width of the exponential function $\gamma$ is activated with a sigmoid function times two, meaning that the initial value before training is $\gamma=1$ when initiating the model weights with zeros.
To verify the predictions of the two proposed parameters $\mu$ and $\gamma$, we conduct an additional experiment in~\cref{suppltab:params}.
The data is collected from GraCE's predictions on the testset of the GroundLink dataset; each type of motion is the average of at least three valid trials, \eg, Ballet is the average result of \texttt{balletsmalljump1, balletsmalljump3, balletsmalljump4}.
Dynamic motions tend to have higher CoM offset due to their fast movement speed, \eg, walking needs $12.64$\unit{\centi\metre} average CoM offset along the moving direction of y-axis.

\begin{table}[ht]
    \centering
    \scriptsize
    \caption{
    The averages and standard deviations of GraCE's gravity-guided parameters $\mu$ and $\gamma$.
    The offset $\mu$ (in \unit{\centi\metre}) is presented in $\mu_{x}$ and $\mu_{y}$ in respective directions.
    \textbf{Bold} and \underline{underline} denote the values with largest and second largest magnitudes.
    }
    \begin{tabular}{l|rrrrrrrrrrrr}
        \toprule
        & Ballet & Hop & Idle & Jump & Dance & Stretch & Soccer & Squat & Taichi & Tree & Walk & Worrier \\
        \midrule
        $\textup{avg}_{\mu_{x}}$ & $-\underline{6.93}$ & $-5.29$ & $-5.75$ & $-5.57$ & $-6.36$ & $-4.71$ & $-2.77$ & $-5.51$ & $-6.82$ & $-4.67$ & $-0.07$ & $\mathbf{-7.88}$ \\
        $\textup{std}_{\mu_{x}}$ & $0.92$ & $1.19$ & $0.34$ & $1.00$ & $0.58$ & $2.04$ & $1.10$ & $0.43$ & $1.62$ & $0.66$ & $2.72$ & $1.03$ \\
        \midrule
        $\textup{avg}_{\mu_{y}}$ & $9.77$ & $0.58$ & $6.82$ & $9.79$ & $10.22$ & $4.18$ & $\underline{10.78}$ & $5.27$ & $4.26$ & $7.95$ & $\mathbf{12.64}$ & $-1.66$ \\
        $\textup{std}_{\mu_{y}}$ & $0.58$ & $0.97$ & $1.59$ & $0.79$ & $1.51$ & $5.41$ & $1.04$ & $4.11$ & $4.14$ & $0.82$ & $4.59$ & $4.45$ \\
        \midrule
        $\textup{avg}_{\gamma}$ & $0.16$ & $\mathbf{0.41}$ & $0.24$ & $0.21$ & $0.18$ & $\underline{0.37}$ & $0.29$ & $0.30$ & $0.21$ & $0.31$ & $0.34$ & $0.19$ \\ 
        $\textup{std}_{\gamma}$ & $0.02$ & $0.05$ & $0.02$ & $0.03$ & $0.03$ & $0.08$ & $0.05$ & $0.09$ & $0.06$ & $0.04$ & $0.15$ & $0.04$ \\
        \bottomrule
    \end{tabular}
    \label{suppltab:params}
\end{table}

\section{Additional details and discussion}
\label{supplsec:clarifications}

\paragraph{Baselines reproduction.}
We take only the contact prediction models from the referenced methods, \eg GroundLinkNet~(CNN)~\cite{han23_groundlink}, ImDyS~(transformer)~\cite{liu25_imdy}, or the explicit contact models from PhysPT~\cite{zhang24_physpt}, EstimCD~\cite{brubaker09_estimcd} and IPMAN~\cite{tripathi23_ipman}.
The explicit contact models have no training parameters and we directly apply them to the SMPL motions. 

\paragraph{Preservation of subject-specific characteristics.}
We consider two characteristics that influence the CoP and GRF predictions: body shape and mass.
Our method performs contact estimation on gender-specific SMPL meshes, obtained via an optimization process on motion captures, thus preserving the shape characteristics of the input subject, \eg foot width and contact area.
On the other hand, it is currently impossible to reliably estimate specific body mass just from motions, without assuming population-average statistics on mass distribution.
Similar to standard protocol in biomechanics, GraCE's predictions are evaluated on the body-weight-normalized reaction forces.

\paragraph{Generalization to different motions.}
GraCE's performance is bounded to quasi-static motions, where the motions are created by the natural reaction of the human body to gravity influence, \ie the Newtonian mechanics hold.
When encountering highly dynamic motions, such as the push-off pose in competitive running, where the CoM is too far from the foot contacts and the runner internally generates massive unobservable forces to the ground, GraCE could predict the GRFs with magnitudes that are close to normal walking or jogging motions.

\end{document}